\documentclass{article}

\usepackage{arxiv}
\usepackage[utf8]{inputenc}
\usepackage[T1]{fontenc}

\usepackage{url}
\usepackage{booktabs}
\usepackage{amsfonts}
\usepackage{nicefrac}
\usepackage{graphicx}
\graphicspath{{./images_pdf/}}
\usepackage{amssymb,amsthm,amsmath}
\usepackage{paralist}
\usepackage{array}
\usepackage{ragged2e}
\usepackage{rotating}
\usepackage{float}
\usepackage{placeins}
\usepackage{needspace}
\usepackage{appendix}

\usepackage{subcaption}
\usepackage{tabularx}
\usepackage{ltablex}

\usepackage[dvipsnames]{xcolor}
\usepackage{listings}
\usepackage[most]{tcolorbox}
\usepackage{microtype}

\usepackage{hyperref}

\definecolor{codegreen}{rgb}{0,0.6,0}
\definecolor{codegray}{rgb}{0.5,0.5,0.5}
\definecolor{codepurple}{rgb}{0.58,0,0.82}
\definecolor{stringcolor}{rgb}{0.5,0,0}
\definecolor{keycolor}{rgb}{0.0,0.0,0.6}

\lstdefinelanguage{YAML}{
  keywords={true, false, null, yes, no, on, off},
  keywordstyle=\color{codepurple}\bfseries,
  ndkeywords={%
    chunks, input, agents, supernode_output, settings_yaml, weights_path,
    target_shape, order, nbins, clip_limit, mask_alpha, output_filename
  },
  ndkeywordstyle=\color{keycolor},
  identifierstyle=\color{black},
  sensitive=true,
  comment=[l]{\#},
  commentstyle=\color{codegray}\itshape,
  stringstyle=\color{stringcolor},
  morestring=[b]',
  morestring=[b]",
  basicstyle=\scriptsize\ttfamily,
  breaklines=true,
  showstringspaces=false,
  frame=none
}

\hypersetup{
    colorlinks=true,
    linkcolor=blue,
    filecolor=magenta,
    urlcolor=cyan
}

\title{Autonomous Computer Vision Development with Agentic AI}

\author{
  Jin Kim\textsuperscript{1},
  M. Wasil Wahi-Anwar\textsuperscript{1},
  Sangyun Park,
  Shawn Shin\textsuperscript{1},
  John M. Hoffman\textsuperscript{1},
  Matthew S. Brown\textsuperscript{1} \\
  \And
  \textsuperscript{1}Center for Computer Vision and Imaging Biomarkers, University of California, Los Angeles, CA, USA
}

\date{\today}

\begin{document}
\maketitle

\let\thefootnote\relax
\let\thefootnote\arabic

\begin{abstract}
Agentic Artificial Intelligence (AI) systems leveraging Large Language Models (LLMs) exhibit significant potential for complex reasoning, planning, and tool utilization. We demonstrate that a specialized computer vision system can be built autonomously from a natural language prompt using Agentic AI methods. This involved extending SimpleMind (SM), an open-source Cognitive AI environment with configurable tools for medical image analysis, with an LLM-based agent, implemented using OpenManus, to automate the planning (tool configuration) for a particular computer vision task. We provide a proof-of-concept demonstration that an agentic system can interpret a computer vision task prompt, plan a corresponding SimpleMind workflow by decomposing the task and configuring appropriate tools. From the user input prompt, "provide sm (SimpleMind) config for lungs, heart, and ribs segmentation for cxr (chest x-ray)"), the agent LLM was able to generate the plan (tool configuration file in YAML format), and execute SM-Learn (training) and SM-Think (inference) scripts autonomously. The computer vision agent automatically configured, trained, and tested itself on 50 chest x-ray images, achieving mean dice scores of 0.96, 0.82, 0.83, for lungs, heart, and ribs, respectively. This work shows the potential for autonomous planning and tool configuration that has traditionally been performed by a data scientist in the development of computer vision applications.
\end{abstract}


\section{Introduction}
Artificial Intelligence (AI) is increasingly moving towards more autonomous systems, often termed \emph{Agentic AI}. These agents are characterized by their ability to reason, plan, and interact with environments and tools to achieve complex goals with minimal human intervention\cite{masterman2024landscape}. Fueled by advancements in Large Language Models (LLMs), which provide powerful natural language understanding and generation capabilities, agentic frameworks are being developed to orchestrate multi-step processes involving sophisticated logic and external interactions\cite{masterman2024landscape, wang2024survey}. Reasoning and planning are central to these agents, enabling them to decompose high-level objectives, formulate strategies, and adapt to dynamic information\cite{dutta2024towards, huang2024understanding, hao2023reasoning}. The potential impact is significant across various domains, including healthcare, where agents might assist with clinical workflows, therapeutic reasoning, or automating administrative tasks\cite{masterman2024landscape}. Recent survey papers \cite{wang2024survey} describe applications in fields ranging from psychology to industrial automation and robotics, but are very limited in computer vision. This project bridges LLM-based planning and the design of computer vision systems using SimpleMind, a Cognitive AI environment with applications to medical imaging.

\subsection{Agentic AI Frameworks: Concepts and Architectures}
Agentic AI systems represent a paradigm shift towards proactive, goal-oriented entities\cite{masterman2024landscape}. These systems typically integrate several core components: a reasoning engine (often an LLM served efficiently, e.g., via vLLM \cite{kwon2023efficient}), perception mechanisms, memory modules, and action modules enabling tool use or environmental interaction\cite{dutta2024towards}.

Architectures vary widely\cite{masterman2024landscape}, single-agent systems rely on one core LLM\cite{masterman2024landscape, wang2024survey}, while multi-agent systems (MAS) involve collaboration between specialized agents, potentially structured vertically (with a leader) or horizontally\cite{masterman2024landscape, wang2024survey}. Effective MAS often benefit from clear leadership, dynamic team structures, and efficient communication protocols\cite{masterman2024landscape, wang2024survey}.

Key techniques underpinning agent capabilities include:
\begin{itemize}
    \item \textbf{Reasoning:} Methods like Chain-of-Thought (CoT)\cite{yao2023react}, Tree-of-Thought (ToT)\cite{hao2023reasoning}, and frameworks like ReAct (Reason+Act) \cite{yao2023react} aim to structure the LLM's inference process, enabling more complex, multi-step problem-solving and traceability\cite{dutta2024towards, hao2023reasoning}.
    \item \textbf{Planning:} Agents decompose high-level goals into manageable steps\cite{dutta2024towards, huang2024understanding}. Planning can involve selecting among alternatives, incorporating feedback for refinement\cite{dutta2024towards, shinn2023reflexion}, or even using external classical planners\cite{dutta2024towards, huang2024understanding}. Frameworks like Reasoning as Planning (RAP) treat reasoning itself as a planning problem, potentially using the LLM as a world model within search algorithms like MCTS\cite{hao2023reasoning}.
    \item \textbf{Tool Use:} Agents overcome LLM limitations (e.g., static knowledge, lack of execution ability) by interacting with external tools via APIs\cite{masterman2024landscape, jiang2024kg}. This can range from web search and code execution \cite{masterman2024landscape} to interacting with specific software or databases\cite{openmanus2025}. Deciding \textit{when} and \textit{which} tool to use is a key reasoning step\cite{masterman2024landscape, jiang2024kg}. Frameworks like KG-Agent specifically utilize toolboxes to interact with structured data like knowledge graphs\cite{jiang2024kg}.
    \item \textbf{Memory:} External memory systems (short-term buffers, long-term databases/vector stores) are crucial for maintaining context beyond the LLM's window and enabling learning from past experiences\cite{masterman2024landscape, dutta2024towards}. Reflection mechanisms allow agents to synthesize insights from memory to guide future actions\cite{shinn2023reflexion, dutta2024towards}.
    \item \textbf{Self-Correction/Adaptation:} Frameworks like Reflexion \cite{shinn2023reflexion} or SALA \cite{shinn2023reflexion} allow agents to learn from failures by reflecting on past actions and generating feedback or corrections to improve subsequent attempts, potentially using a single LLM for both action and adaptation\cite{dutta2024towards}.
\end{itemize}

The OpenManus codebase, used in this work, aligns with these concepts. It represents an asynchronous agent framework using a configurable LLM backend (supporting efficient serving engines like vLLM \cite{kwon2023efficient}), featuring tools for Python execution, browser automation, and file editing, indicating a flexible architecture for implementing agentic capabilities\cite{openmanus2025}.

\subsection{SimpleMind Cognitive AI Framework}
SimpleMind is an open-source software environment designed to integrate deep learning models with symbolic and spatial reasoning, particularly for medical image analysis\cite{choi2023simplemind}. Its core components facilitate a cognitive approach to complex tasks:
\begin{itemize}
    \item \textbf{Configurable Tools:} SimpleMind provides a library of general purpose, configurable Python-based tools. Tool types include `image\_processing` (e.g., resizing, enhancement), `neural\_net` (for segmentation or classification), `mask\_processing` (e.g., morphological operations), `spatial\_inference` (reasoning based on spatial relationships), and `reasoning` (e.g., decision trees, fuzzy logic)\cite{choi2023simplemind}. In particular the SimpleMind agents support spatial reasoning and logical reasoning for object recognition that complement the language-based reasoning provided the LLM for step-by-step planning.
    \item \textbf{Knowledge Graph:} Tool configuration is represented as a SimpleMind knowledge graph (KG) that reflects both procedural and domain knowledge. The KG specifies the tool workflow as a directed graph. The KG is represented in YAML format and configures tools for a specific task by specifying parameter values and data passing relationships. Tools are organized in the KG into "chunks", that abstract a particular function, and chunks are grouped into "supernodes" related to specific objects or processing stages. Directed links in the KG represent data dependencies, and tools activate when their required inputs become available. 
    \item \textbf{Blackboard:} A central working memory where tools post result messages (typically images or masks) and retrieve results from other tools as input, allowing data flow between tools\cite{choi2023simplemind, hayes1985blackboard}. SimpleMind allows for tagging of messages and retrieval of a message chain from the blackboard. This provides strong support for meta level reasoning and opportunistic agent actions and significantly augments the OpenManus agent framework.
\end{itemize}

A typical SimpleMind workflow for segmentation involves steps like image preprocessing, neural network inference, mask post-processing, and a reasoning step (e.g., decision tree) to select the final mask based on features like size, shape, or location\cite{choi2023simplemind}.

\section{Methodology: Bridging Agentic AI and SimpleMind}

This project proposes to extend SimpleMind with an LLM-based agent to automate tool configuration (planning) for a computer vision task. The agent is implemented using OpenManus, running an LLM served via a local vLLM \cite{kwon2023efficient} instance. This approach enables translation of a natural language task description into a structured SimpleMind KG (tool configuration) in YAML format.

\subsection{Conceptual Workflow}
The envisioned workflow involves the following stages:
\begin{enumerate}
    \item \textbf{Task Input:} The user provides the agentic framework with a high-level task description related to medical image analysis (e.g., "Generate a SimpleMind configuration to segment the trachea and lungs from the chest X-ray at 'cxr.nii.gz' and save the output masks.").
    \item \textbf{Planning (Agentic Framework):} The framework's LLM (served by vLLM \cite{kwon2023efficient}), utilizing reasoning techniques (e.g., CoT, task decomposition \cite{dutta2024towards, huang2024understanding}), breaks the task into a logical sequence of steps aligned with SimpleMind's processing patterns. This plan identifies necessary supernodes (e.g., input image, trachea, lungs), the required chunks within each (e.g.,`image\_processing`, `neural\_net`, `mask\_processing`, `candidate\_select`), and tools to be called within each chunk.
    \item \textbf{Tool Selection \& Parameterization (Agentic Framework):} For each step in the plan, the agent determines the appropriate SimpleMind tools and their parameters. This requires the agent to possess knowledge of SimpleMind's YAML configuration format and available agents/parameters. This knowledge could be provided via system prompts, retrieved from an integrated knowledge base (e.g., using RAG techniques \cite{masterman2024landscape}), or potentially stored in the agent's memory\cite{dutta2024towards, masterman2024landscape}. The agent selects the corresponding SimpleMind tool and determines its configuration. Internal tools, like Python execution in OpenManus, may be used if calculations are needed for parameter values. However, in the current implementation the computer vision tools are provided and executed in the SimpleMind framework, rather than OpenManus calling them directly. The OpenManus agent executes SimpleMind modules via a terminal command which applies the plan and executes the tools.
    \item \textbf{YAML Generation (Agentic Framework):} Based on the plan and selected tools/parameters, the agent generates the corresponding YAML syntax for each chunk and supernode, assembling the complete configuration file. This step addresses the challenge of LLMs manipulating structured data formats\cite{yao2023react, masterman2024landscape}.
    \item \textbf{YAML Plan Output:} The agent outputs the complete, formatted YAML file.
    \item \textbf{Training (SimpleMind): }Based on the YAML configuration file, the SM-Learn module is called to perform deep learning of neural network weights.
    \item \textbf{Inference (SimpleMind):} The YAML configuration file is applied by the SM-Think module to execute the image analysis workflow\cite{choi2023simplemind}.
\end{enumerate}

\begin{figure}[ht]
    \centering
    \includegraphics[width=0.9\textwidth]{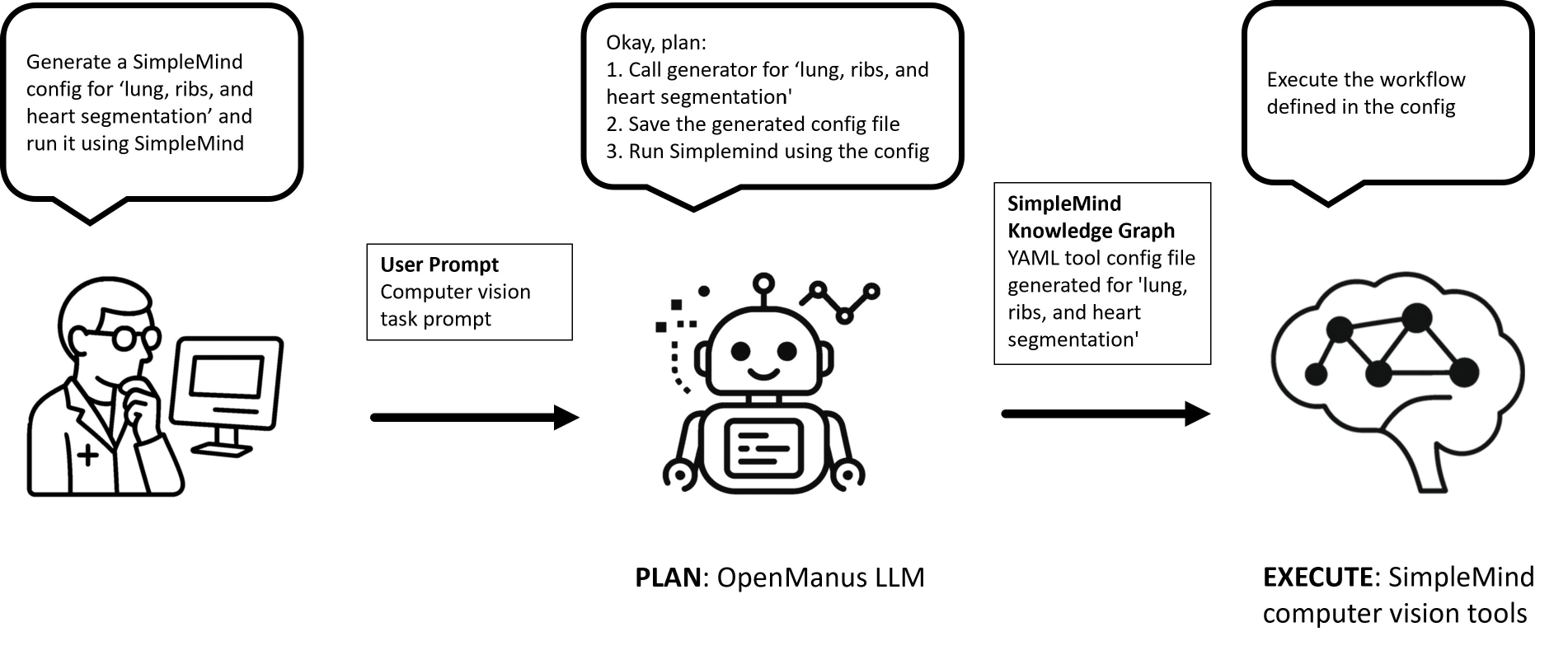} 
    \caption{Conceptual overview of the autonomous workflow. The user provides a natural language prompt (left). The Agentic AI (center-left) plans the task, generates the SimpleMind configuration (center-right), which is then executed by the SimpleMind cognitive AI framework (right) to perform the computer vision task.}
    \label{fig:conceptual_workflow}
\end{figure}

This overall process is depicted in Figure \ref{fig:conceptual_workflow}, illustrating the flow from user prompt to the agentic AI's planning and configuration generation, culminating in the execution by SimpleMind.

\subsection{System Architecture}
\begin{figure}[ht]
    \centering
    \includegraphics[width=\textwidth]{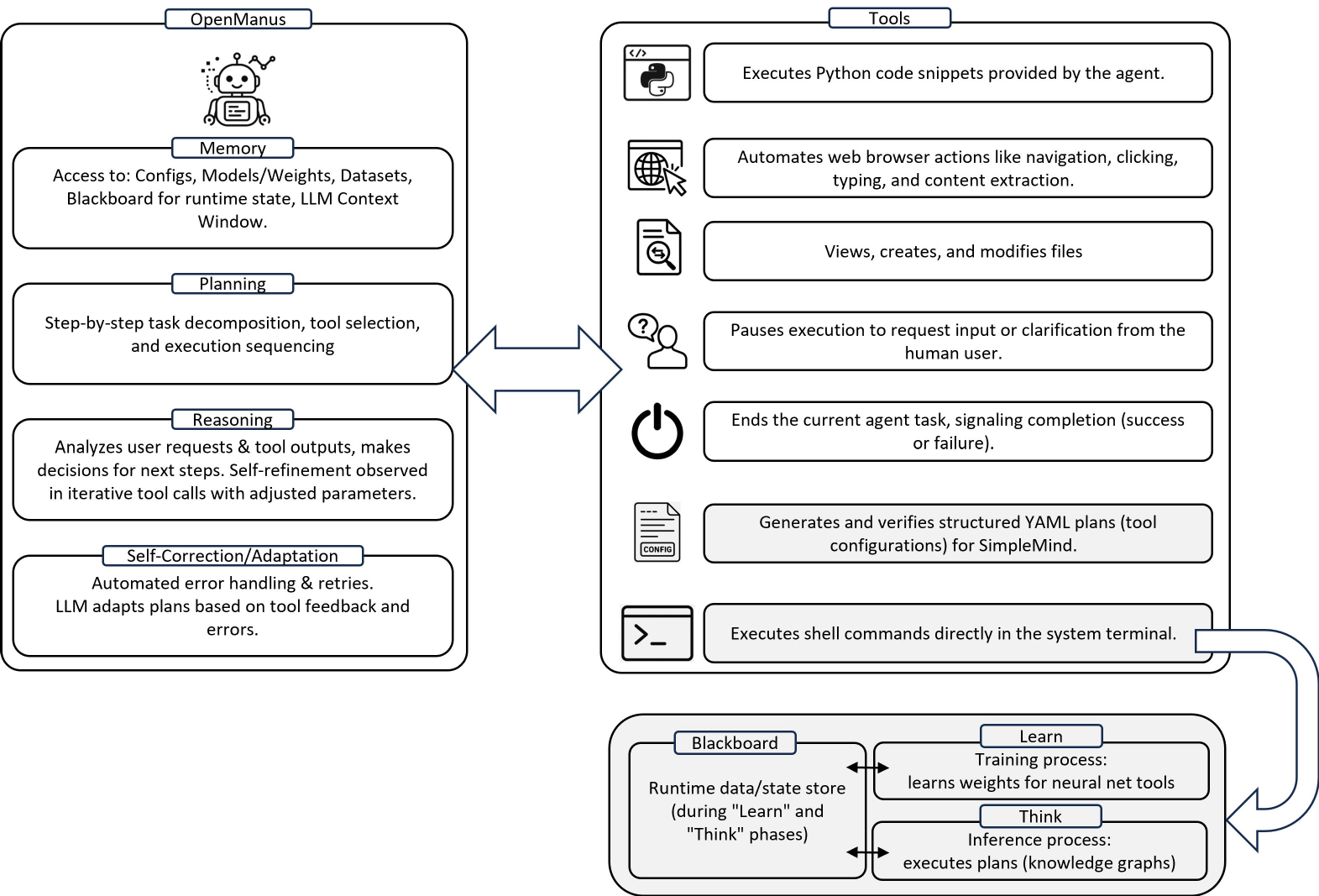} 
    \caption{Detailed system architecture illustrating the interaction between the OpenManus agentic AI framework and the SimpleMind cognitive AI environment. OpenManus (left) components include Memory, Planning, Reasoning, Self-Correction/Adaptation, and various Tools for interacting with the system and generating configurations. It produces a YAML-based Knowledge Graph (right) that defines the workflow for SimpleMind (bottom), which includes a Blackboard for data exchange and processes for Learning and Thinking with the configured computer vision pipeline.}
    \label{fig:system_architecture}
\end{figure}

A detailed schematic diagram of the system architecture, highlighting the interplay between the OpenManus agent and the rest of the SimpleMind environment, is presented in Figure \ref{fig:system_architecture}. The OpenManus agent takes a user prompt, and through a process of reasoning and planning that may involve multiple steps of tool use and self-correction, generates a YAML configuration file. This YAML file defines the knowledge graph for SimpleMind. The SimpleMind framework then executes the computer vision task based on this knowledge graph. SimpleMind components include the Blackboard for passing results between tools, and SM-Learn and SM-Think modules for training CNN models and running inference, respectively.  The initial plan was generated in JSON format by the LLM to minimize formatting issues and automatically converted to YAML format.

\section{Proof of Concept: Chest X-Ray Lungs, Heart, and Ribs Segmentation}

We illustrate the autonomous agentic AI planning with a proof-of-concept example. The task for the computer vision agent was to autonomously generate and execute a plan for lungs, heart, and ribs segmentation on chest X-ray images. 

The initial prompt would include background on the task — documentation on how to construct a SimpleMind knowledge graph, examples of knowledge graphs, and an inventory of computer vision and image processing tools available in SimpleMind. The generated plan should encompass a full pipeline including, for each target anatomy, image preprocessing, deep-learning-based segmentation, and post-processing of the output mask. SimpleMind modules are to be called automatically to handle weight generation (i.e. model training) for the included neural networks and perform inference on the supplied test dataset.

\subsection{Implementation}

\begin{figure}[ht]
    \centering
    \begin{subfigure}[b]{0.5\textwidth}
         \centering
         \includegraphics[width=\textwidth]{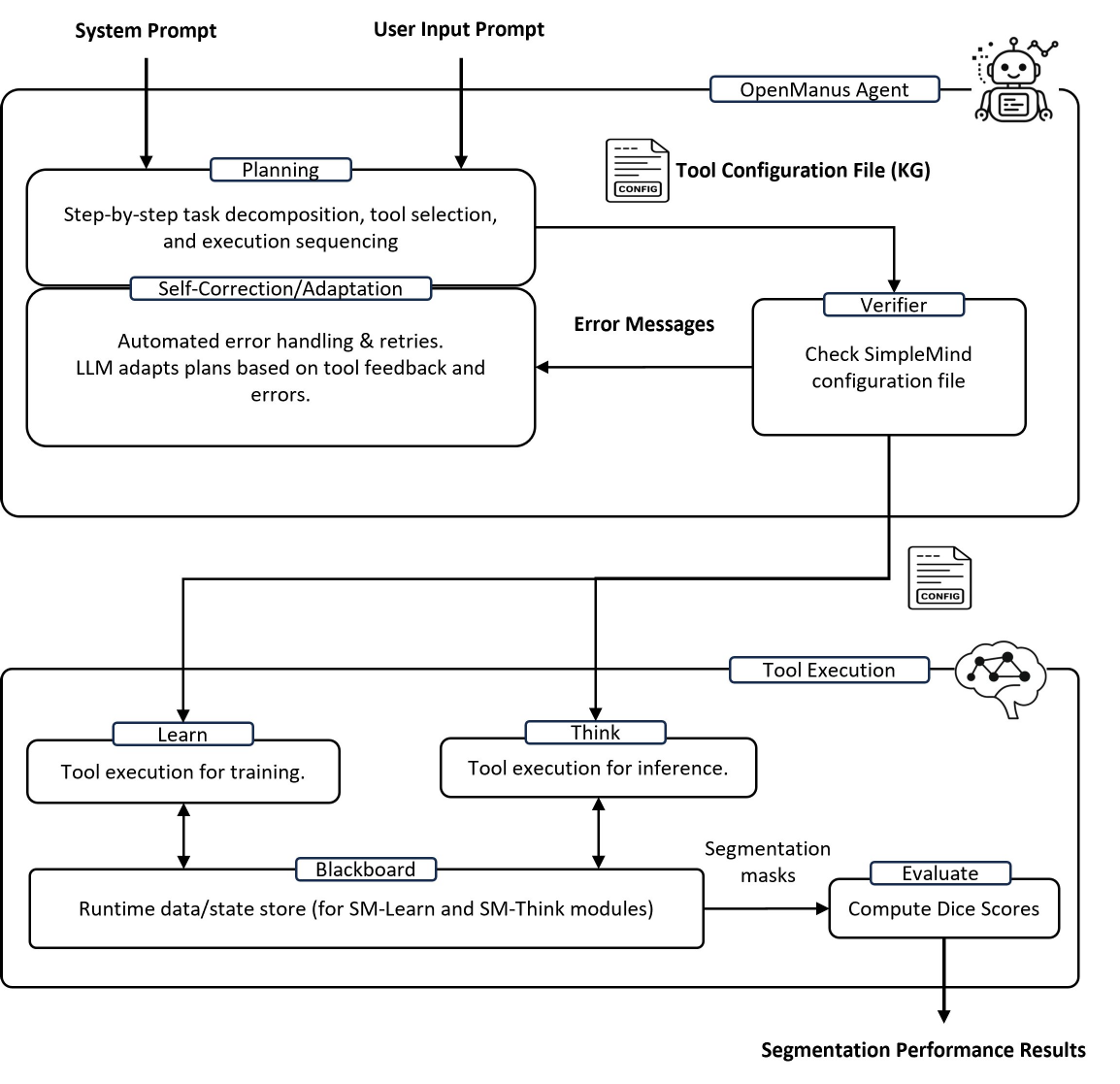}
         \caption{Overall Agentic AI approach, including Planning, Verification, Execution, and Evaluation.}
         \label{fig:fig3_a}
     \end{subfigure}
     \hfill 
     \begin{subfigure}[b]{0.45\textwidth}
         \centering
         \includegraphics[width=\textwidth]{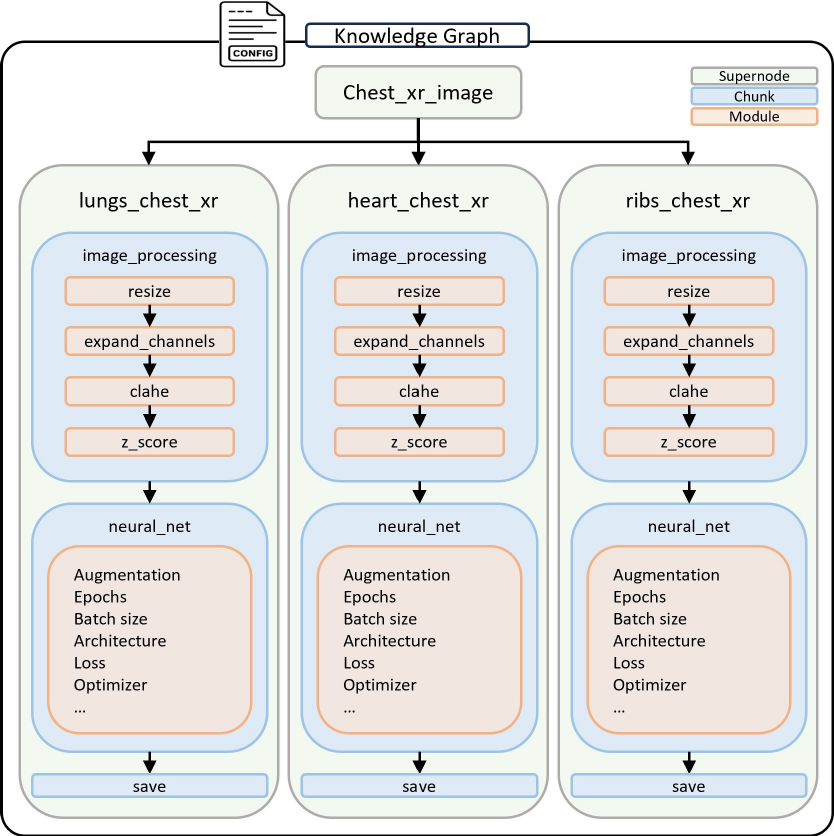}
         \caption{Example SimpleMind Knowledge Graph for Chest X-Ray segmentation.}
         \label{fig:fig3_b}
     \end{subfigure}
    \caption{The Agentic AI approach (a) involves an LLM agent (OpenManus) generating a plan (Knowledge Graph), which is verified before execution. (b) illustrates the structure of the generated Knowledge Graph for segmenting lungs, heart, and ribs, detailing image processing and neural network steps.}
    \label{fig:fig3}
\end{figure}

The overall Agentic AI approach is shown in Figure~\ref{fig:fig3}. System and User input prompts are provided to an LLM agent, implemented using OpenManus. It generates a computer vision plan (in the form of SimpleMind knowledge graph YAML), that is passed to a Verifier to check that the YAML format and tool configurations are valid. The agent invokes the Verifier as a Python script with the YAML file as input. Error messages from the Verifier are provided to the planner as feedback so that the YAML can be iteratively refined.    

Once the YAML is verified, the  knowledge graph is executed by the agent in the SimpleMind environment. First an SM Learn module is called to train the weights for each of the neural network tools, and then an SM Think module performs inference  using the weights on a test set. The inference output will be segmentation masks for the heart, lungs, and ribs, for all cases in the test dataset.

\subsection{Datasets}
In this study, we utilized three separate datasets, each with reference labels (segmentation masks) for a specific anatomical structure: the lungs, heart, and ribs. Each dataset consisted of frontal chest X-ray images paired with their reference segmentation labels. The lung and rib images and annotations were obtained from the Montgomery dataset\cite{candemir2013lung} and VinDr-RibCXR\cite{nguyen2021vindr}, respectively, both of which are publicly available. The heart dataset derives from the COVID dataset\cite{inc2020covid}, with reference labels generated by trained image analysts of the Department of Radiology at the University of California, Los Angeles (UCLA). For model training, we selected the first 100 image-label pairs from each dataset. For testing, we selected the next 50 image-label pairs for the lungs and ribs dataset, and 49 image-label pairs for the heart dataset due to data availability. 

\subsection{User Prompt}
A three-step user prompt was then given to the LLM to instruct it to generate a plan in the form of a SimpleMind knowledge graph (Step 1), and then utilize terminal bash commands to execute the SM Learn training module (Step 2) as well as the SM Think inference module (Step 3).         
\begin{tcolorbox}[
    colback=LimeGreen!10!white,
    colframe=OliveGreen!75!black,
    arc=2mm,
    boxrule=0.5pt,
    left=6mm, right=3mm, top=3mm, bottom=3mm,
    width=\linewidth,
    breakable,  
]
\begin{lstlisting}
1. Provide sm config for lungs, heart, and ribs segmentation for cxr
2. Using terminal, Command "export gpu_num='0' && cd /workdir/simplemind && simplemind/example/train_agents.sh simplemind/example/example.yaml > simplemind/example/learn_output.txt" 
3. Using terminal, "export gpu_num='' && cd /workdir/simplemind && sh simplemind/example/think.sh  > simplemind/example/think_output.txt"

\end{lstlisting}
\end{tcolorbox}

\subsection{Output Plan}
The agent's step-by-step planning process, including interactions with the Verifier and iterative refinement, is detailed in Table \ref{tab:agent_execution}. This table illustrates the agent's thoughts, tool calls, and responses across the multi-step task, culminating in the successful generation and execution of the SimpleMind configuration.

\newcolumntype{P}[1]{>{\RaggedRight\tiny\arraybackslash}p{#1}}
\lstdefinelanguage{JSON}{
    keywords={true,false,null},
    keywordstyle=\color{codepurple}\bfseries,
    stringstyle=\color{stringcolor},
    identifierstyle=\color{black}, 
    commentstyle=\color{codegray}\itshape,
    morestring=[b]", 
    sensitive=true,
    basicstyle=\tiny\ttfamily, 
    breaklines=true,
    showstringspaces=false,
    frame=none,
    literate=
      *{:}{{{\color{black}:}}}{1}
      {,}{{{\color{black},}}}{1}
      {\{}{{{\color{black}\{}}}{1}
      {\}}{{{\color{black}\}}}}{1}
      {[}{{{\color{black}[}}}{1}
      {]}{{{\color{black}]}}}{1}
}
\lstdefinelanguage{Terminal}{ 
    basicstyle=\tiny\ttfamily, 
    breaklines=true,
    showstringspaces=false,
    frame=none,
}
\begin{table*}[htbp]
\centering
\caption{Agent Execution Trace: Generating and Running SimpleMind Configuration.}
\label{tab:agent_execution}

\begin{tabular}{|P{0.31\linewidth}|P{0.31\linewidth}|P{0.31\linewidth}|}
\hline
\multicolumn{3}{|c|}{\textbf{\normalsize User's Multi-Step Request}} \\
\hline
\textbf{\small Task 1: Generate SM Config} & \textbf{\small Task 2: Execute Training} & \textbf{\small Task 3: Execute Thinking} \\
\hline
Provide SM config for lungs, heart, and ribs segmentation for CXR. &
Using terminal, Command: \texttt{export gpu\_num='0' \&\& cd /workdir/simplemind \&\& simplemind/example/train\_agents.sh simplemind/example/example.yaml > simplemind/example/learn\_output.txt} &
Using terminal, Command: \texttt{export gpu\_num='' \&\& cd /workdir/simplemind \&\& sh simplemind/example/think.sh > simplemind/example/think\_output.txt} \\
\hline
\multicolumn{3}{|c|}{\textbf{\normalsize Agent Execution and Refinement Summary}} \\
\hline
\textbf{Initial Thought:} Plan to use \texttt{sm\_json\_generator\_and\_verifier} for config generation.
\vspace{2pt}

\textbf{Tool Call (Initial Attempt for SM Config):}
\begin{lstlisting}[language=JSON]
{
  "name": "sm_json_generator_and_verifier",
  "arguments": {
    "user_request": "lungs, heart, and ribs segmentation for cxr"
  }
}
\end{lstlisting}
\vspace{2pt}
\textbf{Tool Response \& Agent Refinement Loop:}
The initial call with \texttt{user\_request: "lungs, heart..."} prompted the tool, which then failed after 5 internal retries. Key errors encountered during these internal attempts included:
\begin{itemize}[nosep]
    \item JSON extraction failures (Tool Attempts 1, 2, 4 within first agent call).
    \item Verifier: \texttt{chunk\_output} misused as an agent name (Tool Attempt 3).
    \item Verifier: Invalid input linking for \texttt{candidate\_selector} and missing \texttt{decision\_tree} agent (Tool Attempt 3).
    \item Verifier: \texttt{target\_shape} parameter not a stringified list (e.g., \texttt{"[512,512]"}) (Tool Attempt 5).
\end{itemize}
\vspace{1pt}
Recognizing these errors, the agent decided to refine its \texttt{user\_request} for the next call to \texttt{sm\_json\_generator\_and\_verifier}. The refined request explicitly specified the \texttt{target\_shape} format (as \texttt{"'[512,512]'"} ) and emphasized the need for "valid agent inputs".
\vspace{1pt}

A subsequent agent tool call was made with this refined request: \texttt{user\_request: "lungs, heart, ribs segmentation for CXR with resize target\_shape as '[512,512]' and valid agent inputs"}.
\vspace{1pt}

This refined call led to success. The tool internally:
\begin{itemize}[nosep]
    \item Failed JSON extraction on its 1st attempt with the refined request.
    \item Succeeded on its 2nd attempt: JSON syntax validated, YAML conversion successful, and external YAML verification passed ("Checks passed: True").
    \item Final YAML was saved to \texttt{.../example.yaml}.
\end{itemize}
\textit{(Task 1 Complete)}
& 
\textbf{Agent Thought:} SM Config successfully generated. Proceed to execute the training script as per user's step 2.
\vspace{2pt}

\textbf{Tool Call (Training Script):}
\begin{lstlisting}[language=JSON]
{
  "name": "terminal_execute",
  "arguments": {
    "command": "export gpu_num='0' && cd /workdir/simplemind && simplemind/example/train_agents.sh simplemind/example/example.yaml > simplemind/example/learn_output.txt"
  }
}
\end{lstlisting}
\vspace{2pt}
\textbf{Response (Summarized Terminal Output):}
\begin{lstlisting}[language=Terminal]
INFO[0000] Starting `core` server...
INFO[0000] Starting object store on localhost:8082
(*@\dots@*) (*@ Output repeated for 3 cycles @*)
INFO[0234] blackboard http service shutting down
INFO[0234] saving blackboard to blackboard.json
INFO[0234] blackboard shutting down
\end{lstlisting}
The output indicates the script likely executed, with logs saved to \texttt{learn\_output.txt}

\textit{(Task 2 Complete)}
& 
\textbf{Agent Thought:} Training script appears complete. Proceed to execute the thinking script as per user's step 3.
\vspace{2pt}

\textbf{Tool Call (Thinking Script):}
\begin{lstlisting}[language=JSON]
{
  "name": "terminal_execute",
  "arguments": {
    "command": "export gpu_num='' && cd /workdir/simplemind && sh simplemind/example/think.sh > simplemind/example/think_output.txt"
  }
}
\end{lstlisting}
\vspace{2pt}
\textbf{Response (Summarized Terminal Output):}
\begin{lstlisting}[language=Terminal]
INFO[0000] Starting `core` server...
INFO[0000] Starting object store on localhost:8082
INFO[0036] blackboard http service shutting down
INFO[0036] saving blackboard to blackboard.json
INFO[0036] blackboard shutting down
\end{lstlisting}
The output indicates script execution, with logs saved to \texttt{think\_output.txt}.
\vspace{2pt}
\textbf{Agent Final Reasoning:} All three user-requested steps have been executed. The SM config was generated and both terminal commands were run, with outputs indicating completion.
\vspace{2pt}
\textbf{Tool Call (Terminate):}
\begin{lstlisting}[language=JSON]
{
  "name": "terminate",
  "arguments": {"status": "success"}
}
\end{lstlisting}
\textit{(Task 3 Complete \& Interaction End)}
\\ 
\hline
\end{tabular}
\end{table*}

The plan is represented as a SimpleMind knowledge graph in YAML format, that configures computer vision tools into a workflow to accomplish the task. The final output plan was automatically generated after 5 iterations (back-and-forth planning and verification between the two LLM agents previously described). The figure summarizes excerpts of the output for the lungs and heart and gives the full plan for the ribs.

\begin{tcolorbox}[
    colback=LimeGreen!10!white,
    colframe=OliveGreen!75!black,
    arc=2mm,
    boxrule=0.5pt,
    left=6mm, right=3mm, top=3mm, bottom=3mm,
    width=\linewidth,
    breakable,
]
\begin{lstlisting}[language=YAML]

chunks:
  chest_xr_image:
    load_image:
      agents:
        reader:
...

  lungs_chest_xr:
    image_processing:
      input: chest_xr_image
      agents:
        resize:
          target_shape: '[512, 512]'
          order: 3
          preserve_range: true
          anti_aliasing: true
          numpy_only: true
...
    neural_net:
      input_1: chest_xr_image
      input_2: from image_processing
      agents:
        tf2_segmentation:
          prediction_threshold: 0.5
          settings_yaml: simplemind/example/sub/cnn_settings/lungs_cnn_settings.yaml

...

  heart_chest_xr:
    image_processing:
      input: chest_xr_image
...
    neural_net:
      input_1: chest_xr_image
      input_2: from image_processing
...

  ribs_chest_xr:
    image_processing:
      input: chest_xr_image
      agents:
        resize:
          target_shape: '[512, 512]'
          order: 3
          preserve_range: true
          anti_aliasing: true
          numpy_only: true
        expand_channels:
          number_of_channels: 1
          numpy_only: true
        clahe:
          nbins: 256
          clip_limit: 0.03
          channel: 0
          numpy_only: true
        z_score:
          channel: 0
          numpy_only: true
    neural_net:
      input_1: chest_xr_image
      input_2: from image_processing
      agents:
        tf2_segmentation:
          prediction_threshold: 0.5
          settings_yaml: simplemind/example/sub/cnn_settings/ribs_cnn_settings.yaml
          working_dir: ribs_chest_xr
          weights_path: simplemind/example/sub/weights/ribs_chest_xr
          weights_url: https://drive.google.com/file/d/1ribs_id/view?usp=sharing
          supernode_output: true
    save:
      input_1: chest_xr_image
      input_2: ribs_chest_xr
      agents:
        save_image:
          mask_alpha: 0.5
          output_filename: chest_xr_ribs


\end{lstlisting}
\end{tcolorbox}
\begin{figure}[htbp] 
     \centering
     \includegraphics[width=0.9\textwidth]{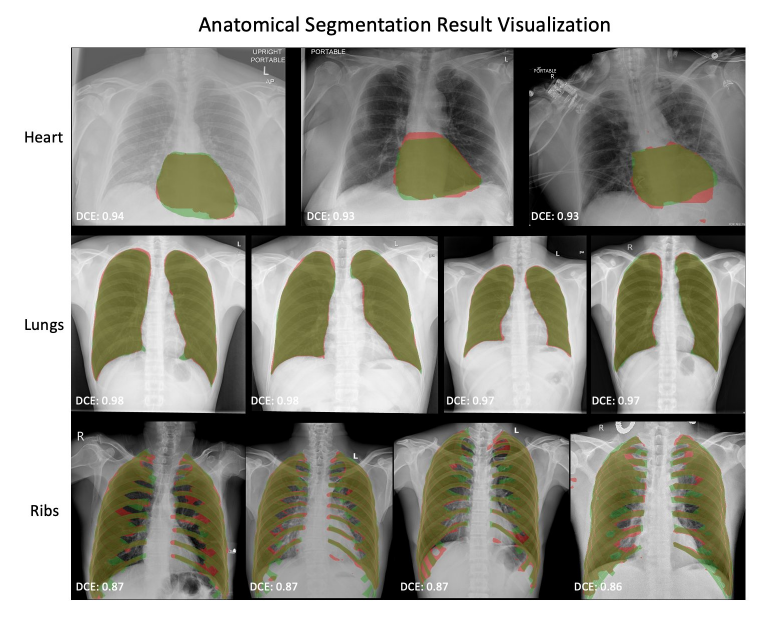} 
     \caption{SimpleMind segmentation result examples for lungs, heart, and ribs in the test set. Red indicates pixels segmented by SimpleMind only (false positive pixels), green are reference pixels only (false negative pixels, missed by the SimpleMind segmentation), and moss green is the overlap (true positive pixels, correctly segmented).}
     \label{fig:segmentation_results}
\end{figure}

\subsection{Segmentation Results}
After generating the knowledge graph, two SimpleMind modules were invoked automatically via a terminal-execution LLM agent: (1) SM-Learn for any tools with trainable weights, and (2) SM-Think for inference using those trained weights. 

On the test set, the system performed with a mean dice coefficient of 0.824 for heart segmentation (49 test cases) , 0.963 for lungs segmentation  (50 test cases) , and 0.830 for ribs    (50 test cases).  Example image segmentation results from the test set are shown in Figure \ref{fig:segmentation_results}.

\subsection{Code Availability}

Instructions for installing the OpenManus LLM agent are provided at https://github.com/jink-ucla/OpenManus-SimpleMind. 

The SimpleMind toolbox used in this manuscript is available as an open-source environment (SimpleMind AI), and available through GitLab (https://gitlab.com/sm-ai-team/simplemind)\cite{choi2023simplemind}. The authors are open to further collaboration and discussion.

Two of the datasets used are available publicly: the Montgomery
dataset for lung segmentation (https://lhncbc.nlm.nih.gov/LHC-downloads/downloads.html), and VinDr-RibCXR for rib segmentation (https://vindr.ai/datasets/ribcxr).

\section{Discussion}
SimpleMind was readily extended with Agentic AI because it was originally developed on the premise that the human mind and thinking processes could be approximated by a configurable multi-agent system. The use of the Blackboard supports this approach.

We successfully demonstrated fully autonomous configuration of a computer vision system. However, the prompting used was quite limited and the LLM likely based its planning largely on provided template examples. The proof of concept example is closely aligned with the online documentation included in the system prompt. The online example includes a tool configuration for trachea segmentation on chest x-ray and partial configuration examples for lungs and heart (ribs are not included). As such, much more extensive experimentation is required to determine the ability of the LLM to generalize planning to other modalities and we expect that more in-depth general documentation will be required to support this. More examples, detailed syntax specification, and more extensive explanation of the available tools will be necessary to handle greater variety of computer vision problems and novel solutions.

This work demonstrates the possibility of replacing the role of the human data scientist in building computer vision applications in SimpleMind, especially when combined with agents to retrieve and format training data sets from the internet, and with existing parameter optimization methods (such as the SimpleMind KnoLO approach \cite{choi2023simplemind}).
It is a work in progress and improvements will follow via our gitlab page and revisions to this manuscript.

\bigskip
\bigskip

\bibliographystyle{unsrt}

\newpage
\appendix
\section{System Prompts}
\subsection{Agent (System) Prompts}
System prompts were supplied to the plan-generating LLM as a preface to the user command. The system prompt began with an introduction.

\begin{tcolorbox}[
    colback=LimeGreen!10!white,
    colframe=OliveGreen!75!black,
    arc=2mm,
    boxrule=0.5pt,
    left=6mm, right=3mm, top=3mm, bottom=3mm,
    width=\linewidth,
]
\begin{lstlisting}
[Prompt Introduction]
Follow the similar pipeline as the example, but you can choose different agents if they function the same. 

Generate a configuration structure in **JSON format** based on the following guidelines and examples. The output should be a single JSON object, starting with a top-level key "chunks". Let's make sure we have the reader at the beginning of the pipeline.

\end{lstlisting}
\end{tcolorbox}

The prompt provided background information to guide the LLM on the structure and contents of the plan, which is to be generated in the form of a SimpleMind knowledge graph (in YAML format). We provide on-line user documentation on the structure of the plan (https://gitlab.com/sm-ai-team/simplemind/-/blob/develop/simplemind/chunks/README.MD?ref\_type=heads). 


\begin{tcolorbox}[
    colback=LimeGreen!10!white,
    colframe=OliveGreen!75!black,
    arc=2mm,
    boxrule=0.5pt,
    left=6mm, right=3mm, top=3mm, bottom=3mm,
    width=\linewidth,
]
\begin{lstlisting}

[JSON Structure Guidelines for Pipeline Configuration]

1. **Hierarchy:** The JSON object has a top-level key `"chunks"`. The value of `"chunks"` is an object containing one or more *supernode* objects. Each *supernode* object is keyed by its unique `supernode_name` (string).
    * Each *supernode* object contains one or more *chunk* objects. Each *chunk* object is keyed by its `chunk_name` (string, unique within the supernode).
    * Each *chunk* object contains key-value pairs for its inputs (see Rule 4) and an `"agents"` object.
    * The `"agents"` object contains one or more *agent* objects. Each *agent* object is keyed by its `agent_name` (see Rule 5 & 7).
    * Each *agent* object contains key-value pairs representing its specific parameters (see Rule 5) and potentially its input sources (see Rule 6) or special flags (`"supernode_output"`, `"chunk_output"`).
\end{lstlisting}

\end{tcolorbox}

The online documentation also included several example plans, including a complete plan for trachea segmentation that configures tools for preprocessing, neural network, post-processing, and selection of the best segmentation candidate. It also included partial examples of plans for lungs and heart segmentation, only including preprocessing and a basic neural network. 

\begin{tcolorbox}[
    colback=LimeGreen!10!white,
    colframe=OliveGreen!75!black,
    arc=2mm,
    boxrule=0.5pt,
    left=6mm, right=3mm, top=3mm, bottom=3mm,
    width=\linewidth,
    breakable,
]
\begin{lstlisting}
[json example]
{
  "chunks": {
    "chest_xr_image": {
      "load_image": {
        "agents": {
          "reader": {
            "csv_path": "__input_images__",
            "header_params": "__header_params__",
            "supernode_output": true
          },
          "save_image": {
            "output_filename": "chest_xr_input_image"
          }
        }
      }
    },
    "trachea_chest_xr": {
      "image_processing": {
        "input": "chest_xr_image",
        "agents": {
          "resize": {
            "target_shape": "[512, 512]",
            "order": 3,
            "preserve_range": true,
            "anti_aliasing": true,
            "numpy_only": true
          },
          "expand_channels": {
            "number_of_channels": 1,
            "numpy_only": true
          },
          "clahe": {
            "nbins": 256,
            "clip_limit": 0.03,
            "channel": 0,
            "numpy_only": true
          },
          "z_score": {
            "channel": 0,
            "numpy_only": true
          }
        }
      },
\end{lstlisting}
\end{tcolorbox}

An inventory of SimpleMind tools was also provided, detailing the expected input and outputs of each tool. Note that most tools expect specifically either a mask or an image input and tool-specific parameters, and output either a mask or image. A plan cannot execute if there is an input-output mismatch in the processing pipeline. 

\begin{tcolorbox}[
    colback=LimeGreen!10!white,
    colframe=OliveGreen!75!black,
    arc=2mm,
    boxrule=0.5pt,
    left=6mm, right=3mm, top=3mm, bottom=3mm,
    width=\linewidth,
    breakable,
]
\begin{lstlisting}

[json dictionary]
{
  "clahe": {
    "info": {
      "agent_input_def": {
        "input": {
          "alternate_names": [
            "image"
          ],
          "optional": false,
          "type": "image_compressed_numpy"
        }
      },
      "agent_output_def": {
        "image": {
          "type": "image_compressed_numpy"
        }
      },
      "agent_parameter_def": {
        "channel": {
          "optional": true
        },
        "clip_limit": {
          "optional": false
        },
        "nbins": {
          "optional": false
        },
        "numpy_only": {
          "optional": false
        }
      }
    },
    "path": "simplemind/agent/image_processing/clahe.py"

  "histeq": {
    "info": {
      "agent_input_def": {
        "input": {
          "alternate_names": [
            "image"
          ],
          "optional": false,
          "type": "image_compressed_numpy"
        }
      },
      "agent_output_def": {
        "image": {
          "type": "image_compressed_numpy"
        }
      },
      "agent_parameter_def": {
        "channel": {
          "optional": false
        },
        "nbins": {
          "optional": false
        },
        "numpy_only": {
          "optional": false
        }
      }
    },
    "path": "simplemind/agent/image_processing/histeq.py"
  },


\end{lstlisting}
\end{tcolorbox}

Finally, the system prompt prefaced the user prompt with this statement:
\begin{tcolorbox}[
    colback=LimeGreen!10!white,
    colframe=OliveGreen!75!black,
    arc=2mm,
    boxrule=0.5pt,
    left=6mm, right=3mm, top=3mm, bottom=3mm,
    width=\linewidth,
]
\begin{lstlisting}
    Looking at the json guideline, json example, and json dictionary,
    please generate an example for:
\end{lstlisting}
\end{tcolorbox}

\end{document}